\documentclass[10pt,twocolumn,letterpaper]{article}

\usepackage{cvpr}
\usepackage{times}
\usepackage{epsfig}
\usepackage{graphicx}
\usepackage{amsmath}
\usepackage{amssymb}

\usepackage{graphics}
\usepackage{enumitem}
\usepackage{bm}
\usepackage{color}
\usepackage{adjustbox}
\usepackage[innercaption]{sidecap}
\usepackage{upquote}
\usepackage{subfigure}
\usepackage{textcomp}
\usepackage{pifont}

\sidecaptionvpos{figure}{m}

\graphicspath{{./}}

\DeclareGraphicsExtensions{.jpeg,.png,.eps,.pdf}

\DeclareMathOperator*{\argmin}{arg\; min}

\newcommand{\partitle}[1]{\noindent\textbf{#1}}

\newcommand{\cmark}{\ding{51}}%
\newcommand{\xmark}{\ding{55}}%

\definecolor{lightgray}{gray}{0.6}

\usepackage[pagebackref=true,breaklinks=true,letterpaper=true,colorlinks,bookmarks=false]{hyperref}

\cvprfinalcopy 

\def\cvprPaperID{379} 

\ifcvprfinal\fi
\begin{document}

\title{Unified Embedding and Metric Learning for Zero-Exemplar Event Detection}
\author{
Noureldien~Hussein, Efstratios~Gavves, Arnold~W.M.~Smeulders
\\
QUVA~Lab, University~of~Amsterdam
\\
{\tt\small\{nhussein,egavves,a.w.m.smeulders\}@uva.nl}}

\maketitle

\begin{abstract}
Event detection in unconstrained videos is conceived as a content-based video retrieval with two modalities: textual and visual. Given a text describing a novel event, the goal is to rank related videos accordingly. This task is zero-exemplar, no video examples are given to the novel event.
\\
Related works train a bank of concept detectors on external data sources. These detectors predict confidence scores for test videos, which are ranked and retrieved accordingly. In contrast, we learn a joint space in which the visual and textual representations are embedded. The space casts a novel event as a probability of pre-defined events. Also, it learns to measure the distance between an event and its related videos.
\\
Our model is trained end-to-end on publicly available EventNet. When applied to TRECVID Multimedia Event Detection dataset, it outperforms the state-of-the-art by a considerable margin.
\\


\end{abstract}

\section{Introduction}

%
%
%
%

TRECVID Multimedia Event Detection (MED)~\cite{over2013trecvid,over2014trecvid} is a retrieval task for event videos, with the reputation of being realistic. It comes in two flavors: few-exemplar and zero-exemplar, where the latter means that no video example is known to the model. Although expecting a few examples seems reasonable, in practice this implies that the user must already have an index of any possible query, making it very limited. In this paper, we focus on event video search with zero exemplars.

Retrieving videos of never-seen events, such as ``renovating home", without any video exemplar poses several challenges. One challenge is how to bridge the gap between the visual and the textual semantics~\cite{jiang2015bridging,habibian2014composite,habibian2015discovering}. One approach~\cite{jiang2015bridging,mazloom2014conceptlets,chang2015semantic,chang2016dynamic,changthey,lu2016zero} is to learn a dictionary of concept detectors on external data source. Then, scores for test videos are predicted using these detectors. Test videos are then ranked and retrieved accordingly. The inherent weakness of this approach is that the presentation of a test video is reduced to a limited vocabulary from the concept dictionary. Another challenge is how to overcome the domain difference between training and test events. While Semantic Query Generation (SQG)~\cite{jiang2015bridging,chang2016dynamic,changthey,jiang2015fast} mitigates this challenge by extracting keywords from the event query, it does not address how relevant these keywords to the event itself. For example, keyword ``person" is not relevant to event ``car repair" as it is to ``flash mob gathering".


\begin{figure}[t]
\begin{center}
\includegraphics[trim=4mm 15mm 6mm 20mm,width=1.0\linewidth]{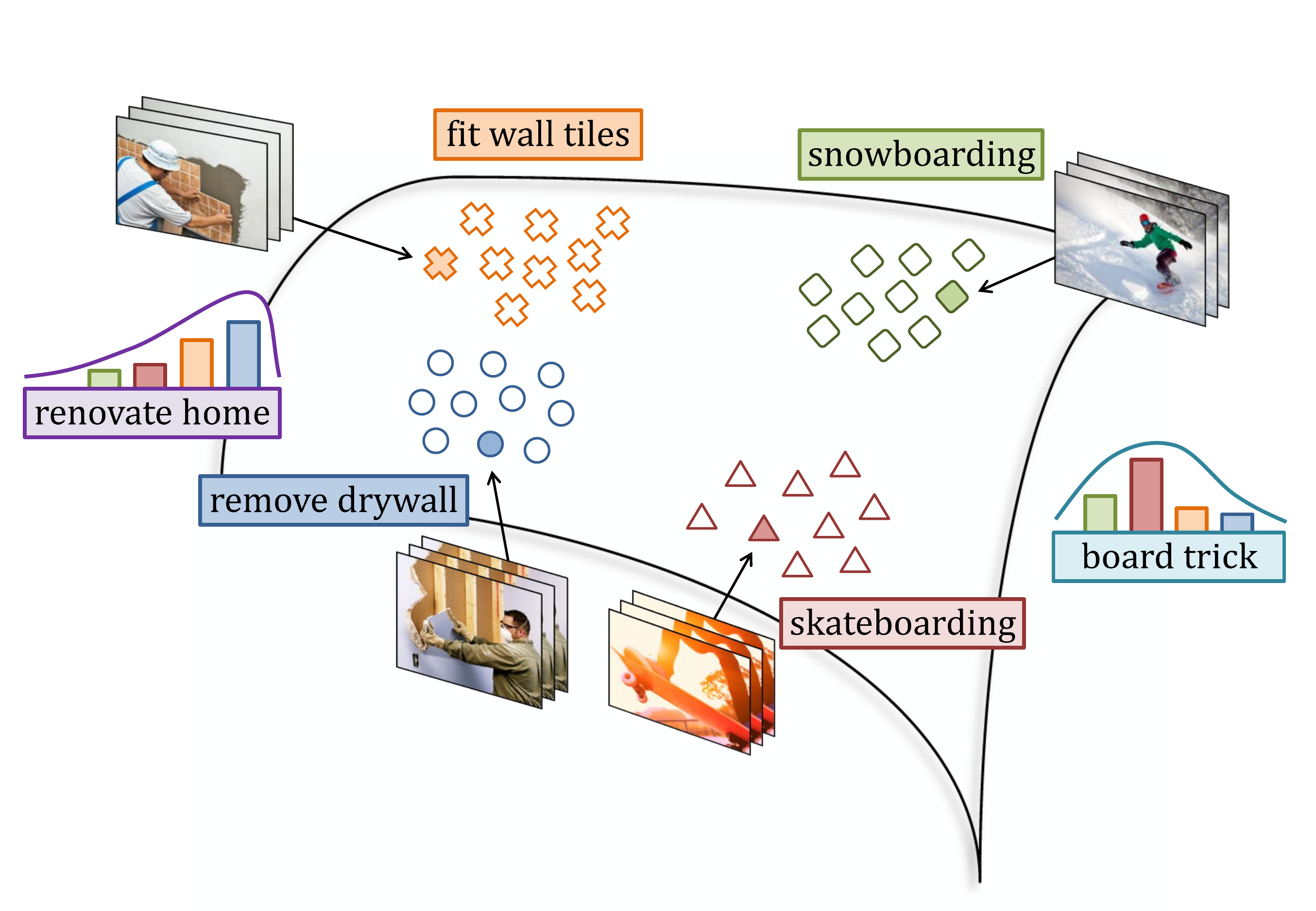}
\end{center}
\caption{We pose the problem of zero-exemplar event detection as learning from a repository of pre-defined events. Given video exemplars of events ``removing drywall" or ``fit wall times", one may detect a novel event ``renovate home" as a probability distribution over the predefined events.}
\label{fig:1-1}
\end{figure}

Our entry to zero-exemplar events is that they generally have strong semantic correlations~\cite{mensink2014costa, gavves2015activetransferlearning} with other possibly seen events. For instance, the novel event ``renovating home" is related to ``fit wall tiles", ``remove drywall", or even to ``paint door". Novel events can, therefore, be casted on a repository of prior events, for which knowledge sources in various forms are available beforehand, such as the videos, as in EventNet~\cite{ye2015eventnet}, or articles, as in WikiHow~\cite{wikihow}. Not only do these sources provide video examples of a large --but still limited-- set of events, but also they provide an association of text description of events with their corresponding videos. A text article can describe the event in words: what is it about, what are the details and what are the semantics. We note that such a visual-textual repository of events may serve as a knowledge source, by which we can interpret novel event queries.

For Zero-exemplar Event Detection (ZED), we propose a neural model with the following novelties:
\begin{enumerate}[leftmargin=1em,itemindent=0em,nolistsep]
\item We formulate a unified embedding for multiple modalities (\eg visual and textual) that enables a contrastive metric for maximum discrimination between events.

\item A textual embedding poses the representation of a novel event as a probability of predefined events, such that it spans a much larger space of admissible expressions.

\item We exploit a single data source, comprising pairs of event articles and related videos. A single source rather enables end-to-end learning from multi-modal individual pairs.

\end{enumerate}
\vspace*{\baselineskip}

We empirically shows that our novelties result in performance improvement. We evaluate the model on TRECVID Multimedia Event Detection (MED) 2013~\cite{over2013trecvid} and 2014~\cite{over2014trecvid}. Our results show significant improvement over the state-of-the-art.


\section{Related Work}\label{sec:related}


\begin{figure}[!ht]
\begin{center}
\includegraphics[trim=0mm 5mm 0mm 10mm,width=0.8\linewidth]{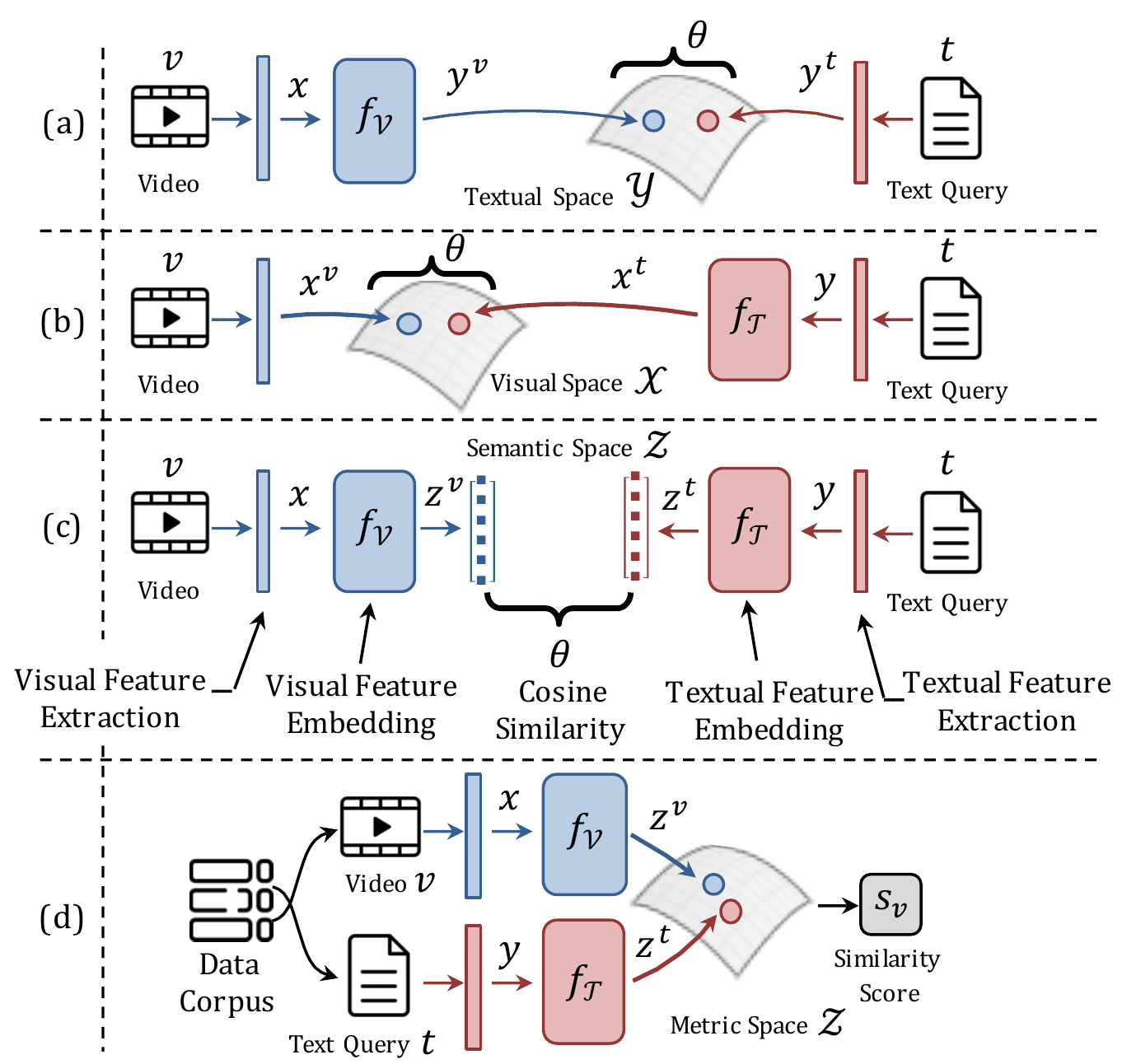}
\end{center}
\caption{Three families of methods for zero-exemplar event detection: (a), (b) and (c). They build on top of feature representations learned a priori (i.e. initial representations), such as CNN features $x$ for a video $v$ or word2vec features $y$ for event text query $t$. In a post-processing step, the distance $\theta$ is measured between the embedded features. In contrast, our model rather falls in a new family, depicted in (d), for it learns unified embedding with metric loss using single data source.}
\label{fig:2-1}
\end{figure}

We identify three families of methods for ZED, as in figure~\ref{fig:2-1} (a), (b) and (c).

\partitle{Visual Embedding and Textual Retrieval.}
As in figure~\ref{fig:2-1}(a), given a video $v_i$ represented as $x \in \mathcal{X}$ and a related text $t$ represented as $y \in \mathcal{Y}$. Then, a visual model $f_\mathcal{V}$ is trained to project $x$ as $y^v \in \mathcal{Y}$ such that the distance is minimized between $\left(y^v, y\right)$. In test time, video ranking and retrieval is done using distance metric between the projected test video $y^t$ and test query representation $y$.

~\cite{habibian2014videostory, habibian2015videostory} project the visual feature $x$ of a web video $v$ into term-vector representation $y$ of the video's textual title $t$. However, during training, the model makes use of the text query of the test events to learn better term-vector representation. Consequently, this limits the generalization for novel event queries.

\partitle{Textual Embedding and Visual Retrieval.}
As in figure~\ref{fig:2-1}(b), a given text query $t$ is projected into $x^t \in \mathcal{X}$ using pre-trained or learned language model $f_\mathcal{T}$.

~\cite{mazloom2015tagbook} makes use of freely-available weekly-tagged web videos. Then it propagates tags to test videos from its nearest neighbors. Methods~\cite{chang2015semantic,chang2016dynamic,changthey,lu2016zero, jiang2015bridging} have similar approach. Given a text query $t$, Semantic Query Generation (SQG) extracts $N$ most related concepts $\left\{c_i, i \in N \right\}$ to the test query. Then, pre-trained concept detectors predict probability scores $\left\{s_i, i \in N \right\}$ for a test video $v$. Aggregating these probabilities results in the final video score $s_v$, upon which videos are ranked and retrieved.~\cite{changthey} learns weighted averaging.

The shortcoming of this family is that expressing a video as probability scores of few concepts is under-representation. Any concept that exists in the video but is missing in the concept dictionary is thus unrepresented.

\partitle{Visual-Textual Embedding and Semantic Retrieval.}
As in figure~\ref{fig:2-1}(c), visual $f_\mathcal{V}$ and textual $f_\mathcal{T}$ models are trained to project both of the visual $x$ and textual $y$ features into a semantic space $\mathcal{Z}$. During test, ranking score is the distance between the projections $z^v, z^t$ in the semantic space $\mathcal{Z}$.

~\cite{wu2014zero} projects video concepts into a high-dimensional lexicon space. Separately, it projects concept-based features to the space, which overcomes the lexicon mismatch between the query and the video concepts. ~\cite{elhoseiny2015zero} embeds a fusion of low and mid-level visual features into distributional semantic manifold~\cite{mikolov2013exploiting,mikolov2013distributed}. In a separate step, it embeds text-based concepts into the manifold.

\begin{SCfigure*}
\includegraphics[trim=-5mm 2mm 2mm 2mm,width=1.3\linewidth]{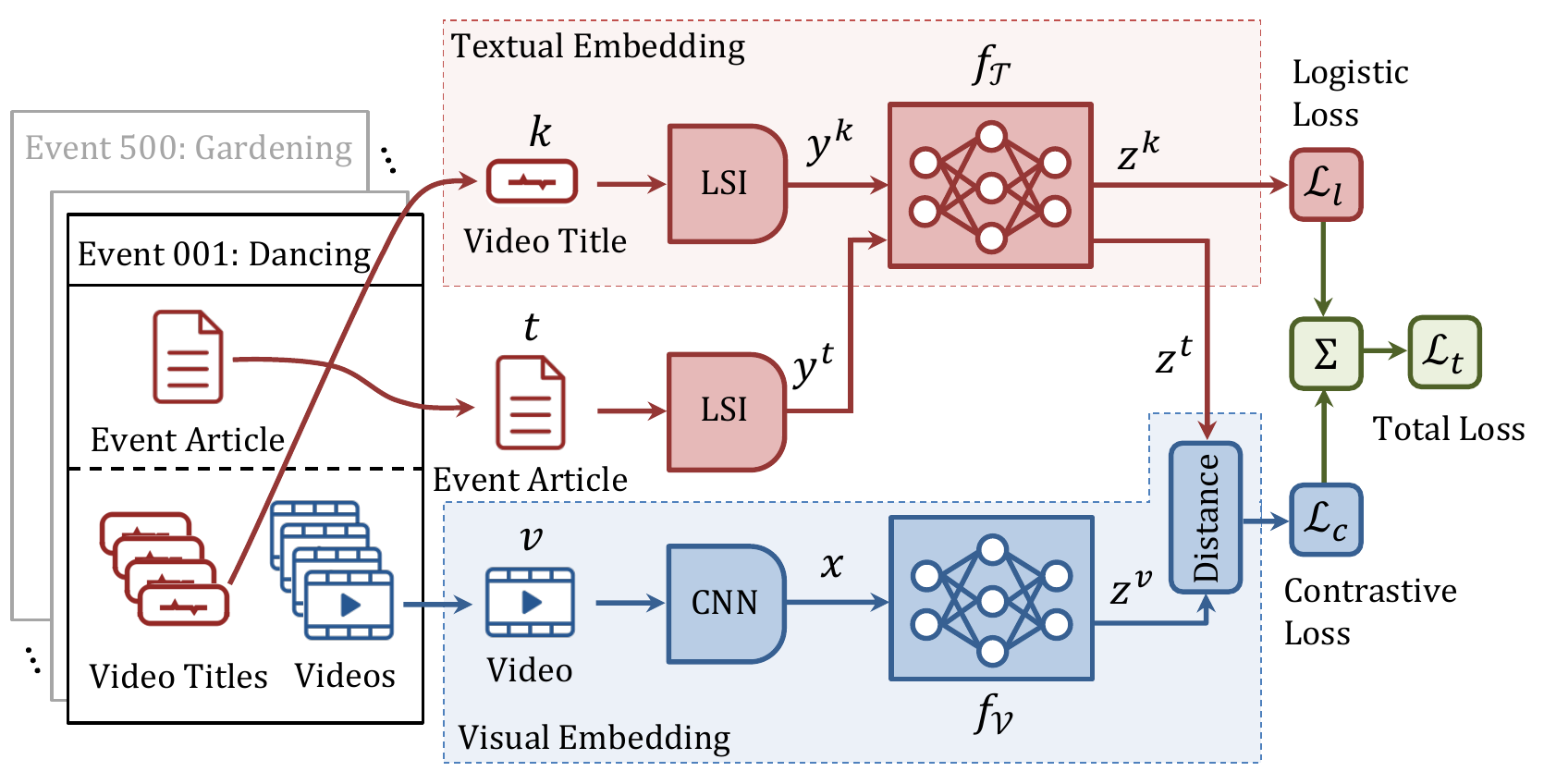}
\caption{Model overview. Using dataset $\mathcal{D}^{z}$ of $M$ event categories and $N$ videos. Each event has a text article and a few videos. Given a video $x$ with text title $k$, belonging to an event with article $t$, we \textbf{extract} features $x, y^k, y^t$ respectively. At the top, network $f_\mathcal{T}$ \textbf{learns} to classify the title feature $y^k$ into one of $M$ event categories. In the middle, we borrow the network $f_\mathcal{T}$ to \textbf{embed} the event article's feature $y^t$ as $z^t \in \mathcal{Z}$. Then, at the bottom, the network $f_\mathcal{V}$ \textbf{learns} to embed the video feature $x$ as $z^v \in \mathcal{Z}$ such that the distance between $\left(z^v, z^t\right)$ is minimized, in the learned metric space $\mathcal{Z}$.}
\label{fig:3-1}
\end{SCfigure*}


The third family, see figure~\ref{fig:2-1}(c), is superior to the others, see figure~\ref{fig:2-1}(a), (b). However, one drawback of~\cite{wu2014zero,elhoseiny2015zero} is separately embedding both the visual and textual features $z^v, z^t$. This leads to another drawback, having to measure the distance between $\left(z^v, z^t\right)$ in a post-processing step (\eg cosine similarity).

\partitle{Unified Embedding and Metric Learning Retrieval}
Our method rather falls into a new family, see figure~\ref{fig:2-1}(d), and it overcomes the shortcomings of~\cite{wu2014zero,elhoseiny2015zero} by the following. It is trained on a single data source, enabling a unified embedding for features of multiple modalities into a metric space. Consequently, the distance between the embedded features is measured by the model using the learned metric space.


\bigbreak
\partitle{Auxiliary Methods}
\label{sub:auxiliary}
Independent to the previous works, the following techniques have been used to improve the results: self-paced reranking~\cite{jiang2014easy}, pseudo-relevance feedback~\cite{jiang2014zero}, event query manual intervention~\cite{agharwal2016tag}, early fusion of features (action~\cite{tran2014c3d,simonyan2014two,wang2011action,uijlings2015video, fernandoTPAMIrankpooling} or acoustic~\cite{muda2010voice,kumar2016audio, jing2016discriminative}) or late fusion of concept scores~\cite{habibian2015videostory}. All these contributions may be applied to our method.

\textbf{Visual Representation.} ConvNets~\cite{lecun1998gradient,he2015deep,krizhevsky2012imagenet,simonyan2014very} provide frame-level representation. To tame them into video-level counterpart, literature use: i- frame-level filtering~\cite{gan2016you} ii- vector encoding~\cite{simonyan2013fisher,arandjelovic2013all} iii- learned pooling and recounting ~\cite{lu2016zero, mettes2015bag} iv- average pooling~\cite{habibian2014videostory,habibian2015videostory}. Also, low-level action~\cite{wang2011action,uijlings2015video}, mid-level action~\cite{tran2014c3d,simonyan2014two} or acoustic~\cite{muda2010voice,kumar2016audio,jing2016discriminative} features can be used.
\textbf{Textual Representation.} To represent text, literature use: i- sequential models~\cite{sutskever2014sequence} ii- continuous word-space representations~\cite{mikolov2013distributed,le2014distributed} iii- topic models~\cite{blei2003latent,deerwester1990indexing} iv- dictionary-space representation~\cite{habibian2015videostory}.

\section{Method}\label{sec:model}
\subsection{Overview}

Our goal is zero-exemplar retrieval of event videos with respect to their relevance to a novel textual description of an event. More specifically, for the zero-exemplar video dataset $\mathcal{D}^{z}=\{v^{z}_i\}, i=1, \dots, L$ and given any future, textual event description $t^{z}$, we want to learn a model $f(\cdot)$ that ranks the videos $v^{z}_i$ according to the relevance to $t^{z}$, namely:
\begin{equation}
t^{z}: v^{z}_i \succ v^{z}_j \rightarrow f(v^{z}_i, t^{z}) > f(v^{z}_j, t^{z}).
\label{eq:def}
\end{equation}


\subsection{Model}

Since we focus on zero-exemplar setting, we cannot expect any training data directly relevant to the test queries.
As such, we cannot directly optimize our model for the parameters $W_\mathcal{T}, W_\mathcal{V}$ in eq.~\eqref{eq:loss_distance_u}. In the absence of any direct data, we resort to external knowledge databases. More specifically, we propose to cast future novel query descriptions as a convex combination of known query descriptions in external databases, where we can measure their relevance the database videos.

We start from a dataset $\mathcal{D}^{z}=\left\{v_i, k_i, l_j, t_j\right\},\;i=1, \dots, N,\; j=1, \dots, M$ organized by an event taxonomy, where we do not neither expect nor require the events to overlap with any future event queries. The dataset is composed of $M$ events. Each event is associated with a textual, article description of the event, analyzing different aspects of it, such as: (i) the typical appearance of subjects and objects (ii) it's procedures (iii) the steps towards completing task associated with it. The dataset contains in total $N$ videos, with $v_i$ denoting the $i$-th video in the dataset with metadata $k_i$, \eg the title of the video. A video is associated with an event label $l_i$ and the article description $t_i$ of the event it belongs to. Since multiple videos belong to the same event, they share the article description of such event. 

The ultimate goal of our model is zero-exemplar search for event videos. Namely, provided unknown text queries by the user, we want to retrieve those videos that are relevant. We illustrate our proposed model during training in figure~\ref{fig:3-1}. The model is composed of two components, a textual embedding $f_\mathcal{T}(\cdot)$, a visual embedding $f_\mathcal{V}(\cdot)$. Our ultimate goal is the ranking of videos, $v_i \succ v_j \succ v_k$ with respect to their relevance to a query description, or in pairwise terms $v_i \succ v_j$, $v_j \succ v_k$ and $v_i \succ v_k$.

Let us assume a pair of videos $v_i$, $v_j$ and query description $t$, where video $v_i$ is more relevant to the query $t$ than $v_j$. Our goal is a model that learns to put videos in the correct relative order, namely $(v_i, t) \succ (v_j, t)$. This is equivalent to a model that learns visual-textual embeddings such that $d^{tv}_i < d^{tv}_j$, where $d^{tv}_i$ is the distance between visual-textual embeddings of $(v_i, t)$, $d^{tv}_j$ is the same for $(v_j, t)$.
Since we want to compare distances between pairs $(v_i, t), (v_j, t)$, we pose the learning of our model as the minimization of a contrastive loss~\cite{chopra2005learning}:
\begin{align}
\mathcal{L}_{con} &= \frac{1}{2N} \sum_{i=1}^N h_i \cdot d_i^2 + (1-h_i) \max(1 - d_i, 0)^2,
\label{eq:loss_contrastive_u}
\\
d_i &= \|f_\mathcal{T}(t_i; W_\mathcal{T}) - f_\mathcal{V}(v_i; W_\mathcal{V})\|_2,
\label{eq:loss_distance_u}
\end{align}
where $f_\mathcal{T}(t_i; W_\mathcal{T})$ is the projection of the query description $t_i$ into the \textbf{unified} metric space $\mathcal{Z}$ parameterized by $ W_\mathcal{T}$, $f_\mathcal{V}(v_i;W_\mathcal{V})$ is the projection of a video $v_i$ onto the same space $\mathcal{Z}$ parameterized by $ W_\mathcal{V}$ and $h_i$ a target variable that equals to $1$ when the $i$-th video is relevant to the query description $t_i$ and $0$ otherwise.
Naturally, to optimize eq.~\eqref{eq:loss_contrastive_u}, we first need to define the projections $f_\mathcal{T}(\cdot; W_\mathcal{T})$ and $f_\mathcal{V}(\cdot; W_\mathcal{V})$ in eq.~\eqref{eq:loss_distance_u}.

\bigbreak
\partitle{Textual Embedding.}
The textual embedding component of our model, $f_\mathcal{T}(\cdot; W_\mathcal{T})$, is illustrated in figure~\ref{fig:3-1} (top). This component is dedicated to learn a projection of a textual input --including any future event queries $t$-- on to the unified space $\mathcal{Z}$. Before detailing our model $f_\mathcal{T}$, however, we note that that the textual embedding can be employed not only with event article descriptions, but also with any other textual information that might be associated to the dataset videos, such as textual metadata. Although we expect the video title not to be as descriptive as the associated article, they may still be able to offer some discriminative information as previously shown~\cite{habibian2014videostory,habibian2015videostory} which can be associated to the event category.

We model the textual embedding as a shallow (two layers) multi-layer perceptron (MLP). For the first layer we employ a \texttt{ReLU} nonlinearity. The second layer serves a dual purpose. First, it projects the article description of an event on the unified space $\mathcal{Z}$. This projection is \emph{category-specific}, namely different videos that belong to the same event will share the projection. Second, it can project any \emph{video-specific} textual metadata into the unified space. We, therefore, propose to embed the title metadata $k_i$, which is uniquely associated with a video, not an event category. To this end, we opt for \texttt{softmax} nonlinearity for the second layer, followed by an additional logistic loss term to penalized misprediction of titles $m_i$ with respect to the video's event label $y_i^j$, namely
\begin{equation}
\mathcal{L}_{log} = \sum_{i=1}^N \sum_{j=1}^M -y_i^j \log f_\mathcal{T}^j(k_i; W_\mathcal{T}).
\label{eq:loss_logistic_u}
\end{equation}

Overall, the textual embedding $f_\mathcal{T}^j$ is trained with a dual loss in mind.
The first loss term, see eq.~\eqref{eq:loss_contrastive_u}~\eqref{eq:loss_distance_u} takes care that the final network learns event-relevant textual projections. The second loss term, see eq.~\eqref{eq:loss_logistic_u}, takes care that the final network does not overfit to the particular event article descriptions. The latter is crucial because the event article descriptions in $\mathcal{D}^{z}$ will not overlap with the future event queries, since we are in a zero-exemplar retrieval setting. As such, training the textual embedding to be optimal only for these event descriptions will likely result in severe overfitting. Our goal and hope is that the final textual embedding model $f_\mathcal{T}$ will capture both event-aware and video-discriminative textual features.


\bigbreak
\partitle{Visual Embedding.}
The visual embedding component of our model, $f_\mathcal{V}(\cdot; W_\mathcal{V})$, is illustrated in figure~\ref{fig:3-1} (bottom). This component is dedicated to learn a projection from the visual input, namely the videos in our zero-exemplar dataset $\mathcal{D}^{z}$, into the unified metric space $\mathcal{Z}$. The goal is to project the videos belonging to semantically similar events; project them into a similar region in the space. We model the visual embedding $f_\mathcal{V}(v_i; W_\mathcal{V})$ using a shallow (two layers) multi-layer perceptron with \texttt{tanh} nonlinearities, applied to any visual feature for video $v_i$.


\bigbreak
\partitle{End-to-End Training.}
At each training forward-pass, the model is given a triplet of data inputs, an event description $t_i$, a related video $v_i$ and video title $k_i$. From eq.~\eqref{eq:loss_distance_u} we observe that the visual embedding $f_\mathcal{V}(v_i; W_\mathcal{V})$ is encouraged to minimize its distance with the output of the textual embedding $f_\mathcal{T}(t_i; W_\mathcal{T})$. In the end, all the modules of the proposed model are differentiable. Therefore, we train our model in an end-to-end manner by minimizing the following objective
\begin{equation}
\begin{aligned}
& \argmin_{W_\mathcal{V}, W_\mathcal{T}} \;\; \mathcal{L}^{\mathcal{U}},
\\
& \mathcal{L}^{\mathcal{U}} = \mathcal{L}_{con} + \mathcal{L}_{log}.
\label{eq:loss_model_u}
\end{aligned}
\end{equation}
For the triplet input $\left(v_i, t_i, k_i\right)$, we rely on external representations, since our ultimate goal is zero-exemplar search. Strictly speaking, a visual input $v_i$ is represented as CNN~\cite{he2015deep} feature vector, while textual inputs $t_i, k_i$ are represented as LSI~\cite{deerwester1990indexing} or Doc2Vec~\cite{le2014distributed} feature vectors. However, given that these external representations rely on neural network architectures, if needed, they could also be further fine-tuned. We choose to freeze CNN and Doc2Vec modules to speed up training. Finally, in this paper we refer to our main model with unified embedding, as \textbf{model}$^\mathcal{U}$.

\bigbreak
\partitle{Inference.}
After training, we fix the parameters ($W_\mathcal{V}, W_\mathcal{T})$. At test time, we set our function $f(\cdot)$ from eq.~\eqref{eq:def} to be equivalent to the distance function from eq.~\eqref{eq:distance}. Hence, at test time, we compute the Euclidean distance in the learned metric space $\mathcal{Z}$ between the embeddings $\left(z^v, z^t\right)$ of test video $v$ and novel event description $t$, respectively.

\section{Experiments}\label{sec:experiments}
\subsection{Datasets}

Before delving into the details of our experiments, first we describe the external knowledge sources we use.

\bigbreak
\partitle{Training dataset.}
We leverage videos and articles from publicly available datasets. EvenNet~\cite{ye2015eventnet} is a dataset of $\sim$90k event videos, harvested from YouTube and categorized into 500 events in hierarchical form according to the events' ontology. Each event category contains around $180$ videos. Each video is coupled with a text title, few tags and related event's ontology.

We exploit the fact that all events in EventNet are harvested from WikiHow~\cite{wikihow} -- a website for \textit{How-To} articles covering a wide spectrum of human activities. For instance: ``How to Feed a Dog" or ``How to Arrange Flowers". Thus, we crawl WikiHow to get the articles related to all the events in EventNet.

\bigbreak
\partitle{Test dataset.}
As the task is zero-exemplar, the test sets are different from the training. While EventNet serves as the training, the following serve as the test: TRECVID MED-13~\cite{over2013trecvid} and MED-14~\cite{over2013trecvid}. In details, they are datasets of videos for events. They comprise 27k videos. There are two versions, MED-13 and MED-14 with 20 events for each. Since 10 events overlap, the result is 30 different events in total. Each event is coupled with short textual description (title and definition).


\subsection{Implementation Details}

\partitle{Video Features.} To represent a video $v$, we uniformly sample a frame every one second. Then, using ResNet~\cite{he2015deep}, we extract \texttt{pool5} CNN features for the sampled frames. Then, we average pool the frame-level features to get the video-level feature $x^v$. We experiment different features from different CNN models: ResNet (\texttt{prob}, \texttt{fc1000}), VGG~\cite{simonyan2014very} (\texttt{fc6}, \texttt{fc7}), GoogLeNet~\cite{szegedy2015going} (\texttt{pool5}, \texttt{fc1024}), and Places365~\cite{zhou2016places} (\texttt{fc6}, \texttt{fc7,fc8}) except we find ResNet \texttt{pool5} to be the best. We only use ResNet \texttt{pool5} and we don't fuse multiple CNN features.

\bigbreak
\partitle{Text Features.} We choose topic modeling~\cite{blei2003latent,deerwester1990indexing}, as it is well-suited for long (and sometimes noisy) text articles. We train LSI topic model~\cite{deerwester1990indexing} on Wikipedia corpus~\cite{wikipedia}. We experiment different latent topics ranging from $300$ to $6000$, expect we found 2500 to be the best. Also, we experiment other textual representations as LDA~\cite{blei2003latent}, SkipThoughts~\cite{kiros2015skip} and Doc2Vec~\cite{le2014distributed}. To extract a feature from an event article $k$ or video title $t$, first we preprocess the text using standard MLP steps: tokenization, lemmatization and stemming. Then, for $k, t$ we extract 2500-D LSI features $y^k, y^t$, respectively. The same steps apply to MED text queries.

\bigbreak
\partitle{Model Details.}
Our visual and textual embeddings $f_{\mathcal{V}}(\cdot), f_{\mathcal{T}}(\cdot)$ are learned on top of the aforementioned visual and textual features ($x^v, y^k, y^t$). $f_\mathcal{T}(\cdot)$ is a 1-hidden layer MLP classifier with \texttt{ReLU} for hidden, \texttt{softmax} for output, \texttt{logistic} loss and \texttt{2500-2500-500} neurons for the input, hidden, and output layers, respectively. Similarly, $f_\mathcal{V}(\cdot)$ is a 1-hidden layer MLP regressor with \texttt{ReLU} for hidden, \texttt{contrastive} loss and \texttt{2048-2048-500} neurons for the input, hidden, and output layers, respectively. Our code is made public\footnote{\url{github.com/noureldien/unified_embedding}} to support further research.

\subsection{Textual Embedding}

\begin{figure}[!ht]
\centering
\subfigure[LSI Features]{
\includegraphics[trim=0mm 0mm 0mm 8mm,scale=0.56]{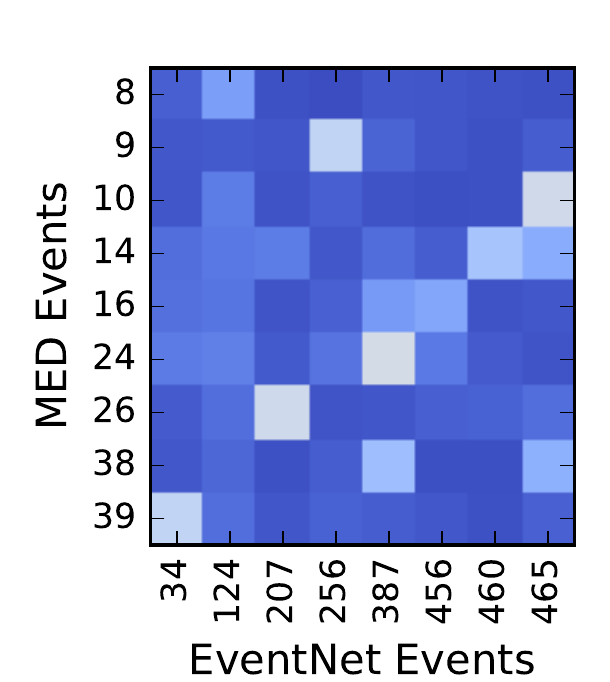}
\label{fig:4-2-1}}
\vspace{0cm}
\subfigure[Embedded Features]{
\includegraphics[trim=0mm 0mm 0mm 8mm,scale =0.56]{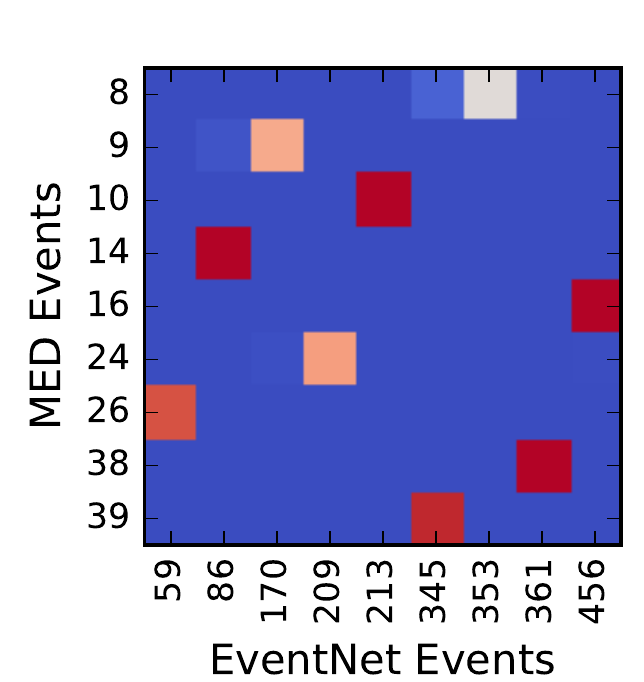}
\label{fig:4-2-2}}
\caption{Our textual embedding (b) maps MED to EventNet events better than LSI features. Each dot in the matrix shows the similarity between MED and EventNet events.}
\label{fig:4-2}
\end{figure}

Here, we qualitatively demonstrate the benefit of the textual embedding $f_\mathcal{T}(\cdot)$. Figure~\ref{fig:4-2} shows the similarity matrix between MED and EventNet events. Each dot represents how a MED event is similar to EventNet events. It shows that our embedding (right) is better than LSI (left) in mapping MED to EventNet events. For example, LSI wrongly maps ``9: getting a vehicle unstuck" to ``256: launch a boat" while our embedding correctly maps it to ``170: drive a car". Also, our embedding maps with higher confidence than LSI, as in ``16: doing homework or study".

\begin{figure}[!ht]
\centering
\subfigure[LSI Features]{
\includegraphics[trim=5mm 0.5mm 0mm 5mm,scale=0.58]{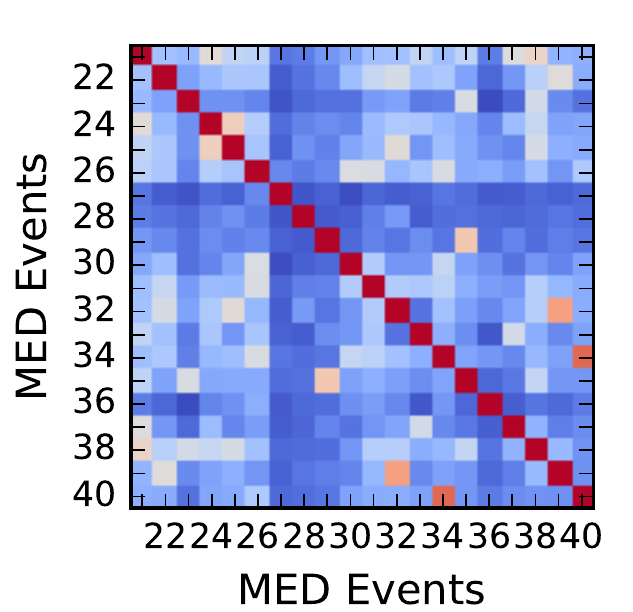}
\label{fig:4-1-1}}
\subfigure[Embedded Features]{
\includegraphics[trim=5mm 0mm 5mm 5mm,scale=0.58]{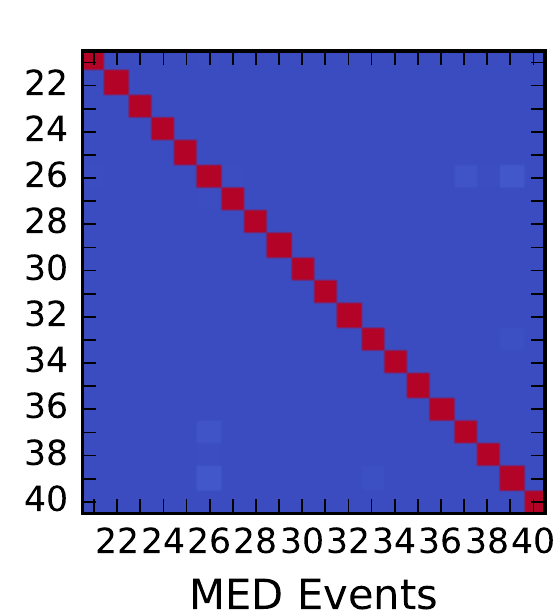}
\label{fig:4-1-2}}
\caption{For 20 events of MED-14, our textual embedding (right) is more discriminant than the LSI feature representation (left). Each dot in the matrix shows how similar an event to all the others.}
\label{fig:4-1}
\end{figure}

Figure~\ref{fig:4-1} shows the similarity matrix for MED events, where each dot represents how related any MED event to all the others. Our textual embedding (right) is more discriminant than on the LSI feature representation (left). For example, LSI representation shows high semantic correlation between events ``34: fixing musical instrument" and ``40: tuning musical instrument", while our embedding discriminate them.


\begin{figure*}[!ht]
\vspace*{\fill}
\centering
\subfigure[MED-13, visual embedding (\textbf{model}$^\mathcal{V}$).]{
\includegraphics[trim=10mm 0mm 10mm 5mm,width=0.3\linewidth]{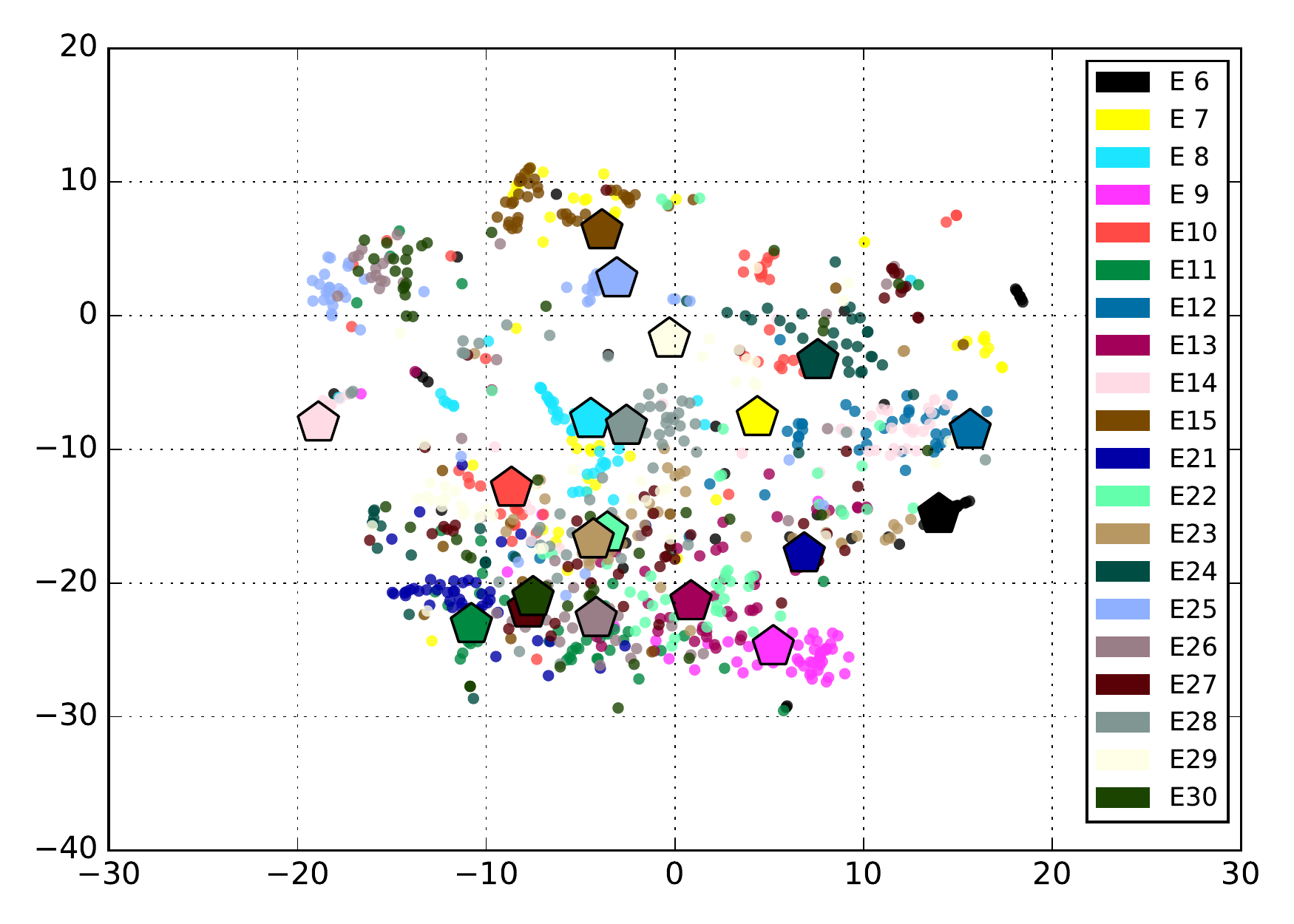}
\label{fig:4-5-1}}
\hfill
\subfigure[MED-13, separate embedding (\textbf{model}$^\mathcal{S}$).]{
\includegraphics[trim=10mm 0mm 10mm 5mm,width=0.3\linewidth]{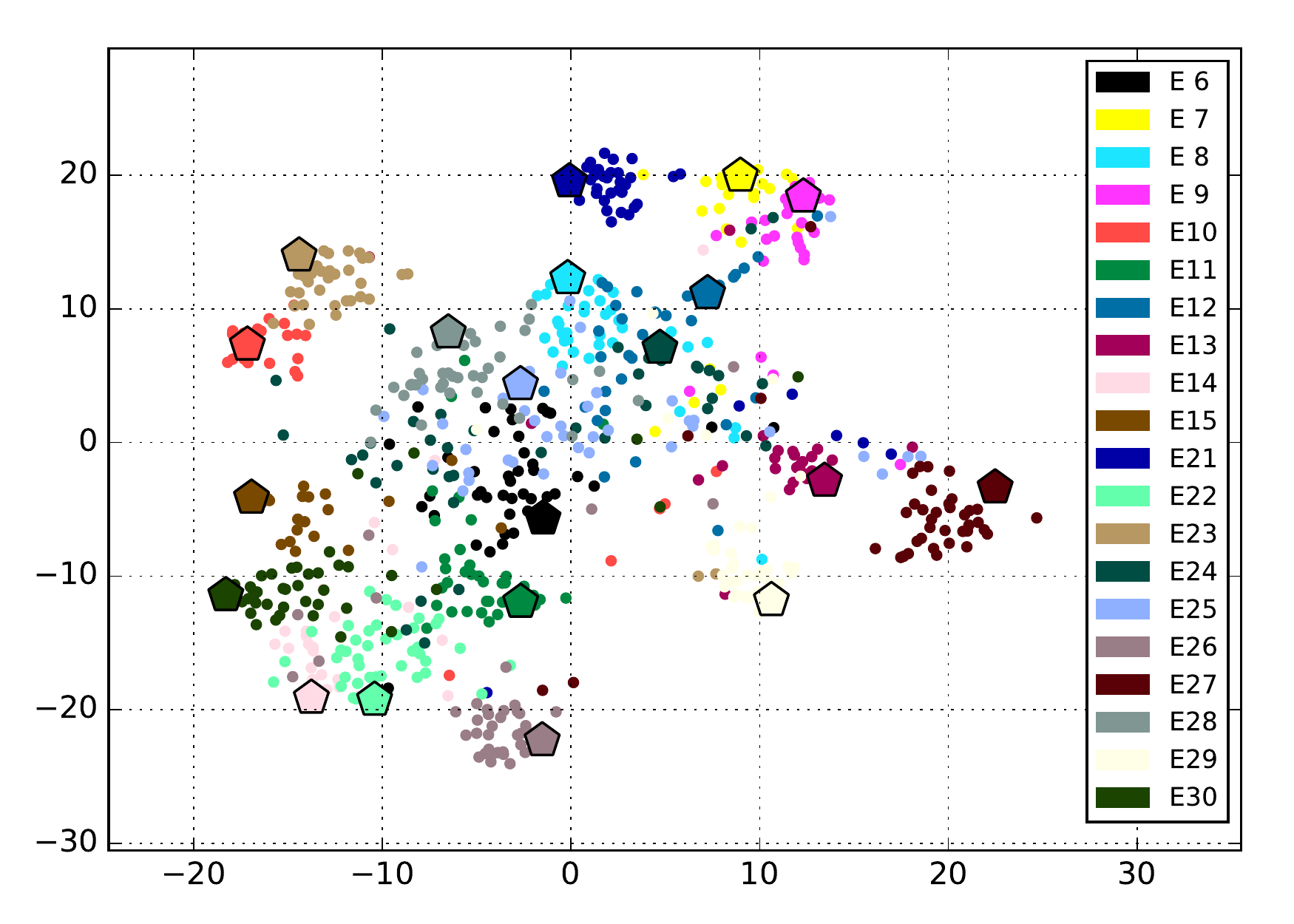}
\label{fig:4-5-2}}
\hfill
\subfigure[MED-13, unified embedding (\textbf{model}$^\mathcal{U}$).]{
\includegraphics[trim=10mm 0mm 10mm 5mm,width=0.3\linewidth]{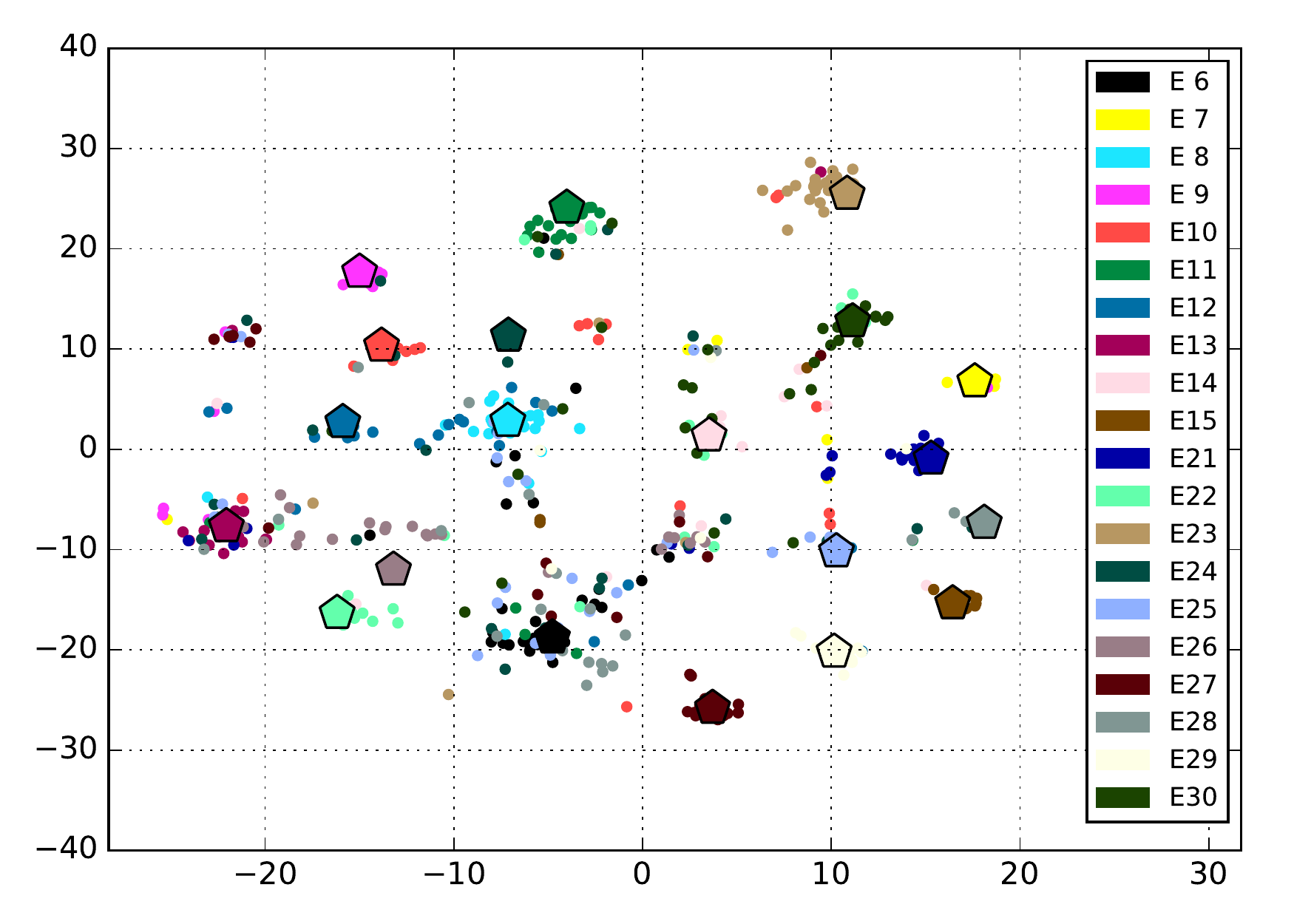}
\label{fig:4-5-3}}
\vskip\baselineskip
\subfigure[MED-14, visual embedding (\textbf{model}$^\mathcal{V}$).]{
\includegraphics[trim=10mm 0mm 10mm 5mm,width=0.3\linewidth]{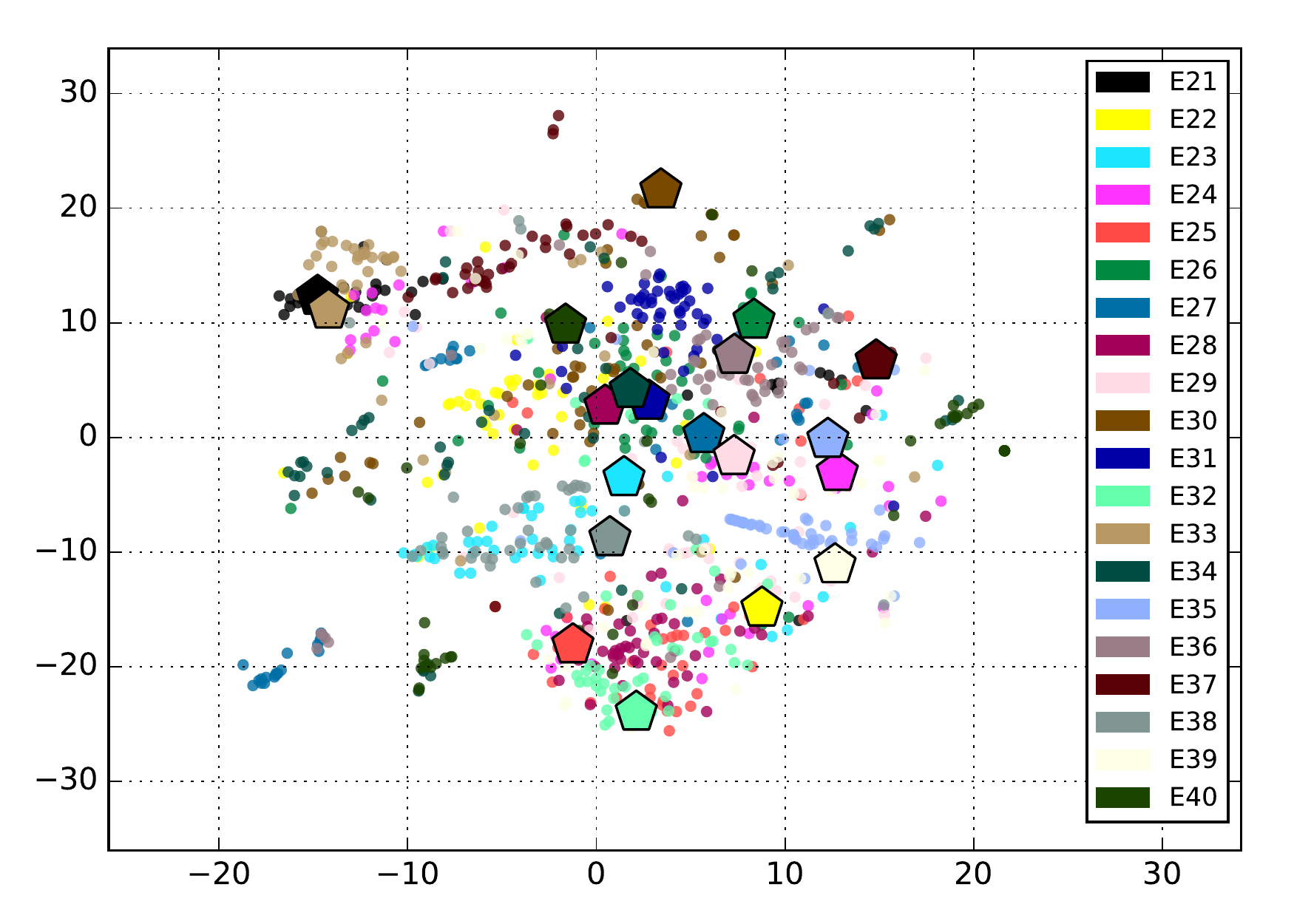}
\label{fig:4-5-4}}
\hfill
\subfigure[MED-14 separate embedding (\textbf{model}$^\mathcal{S}$).]{
\includegraphics[trim=10mm 0mm 10mm 5mm,width=0.3\linewidth]{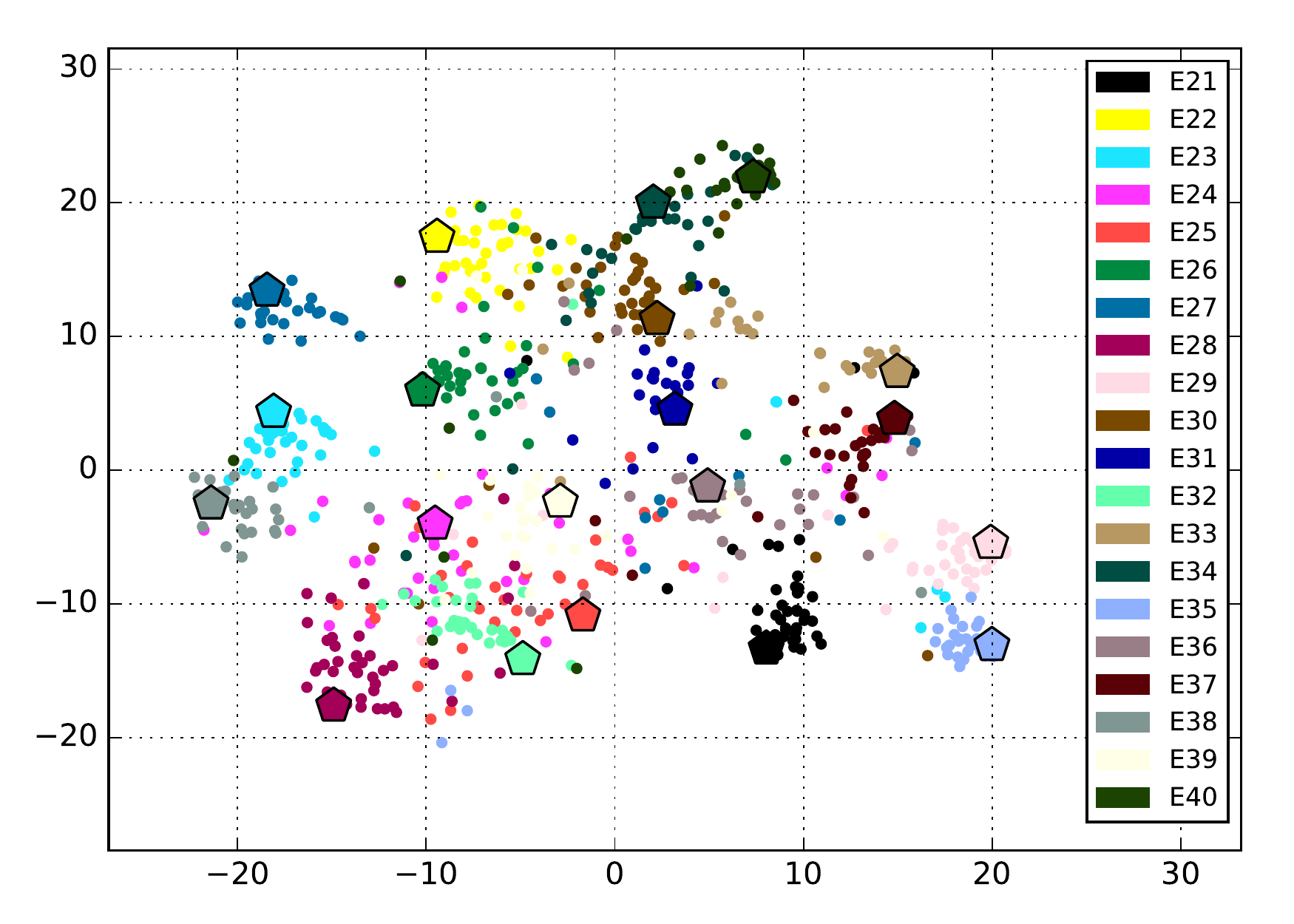}
\label{fig:4-5-5}}
\hfill
\subfigure[MED-14, unified embedding (\textbf{model}$^\mathcal{U}$).]{
\includegraphics[trim=10mm 0mm 10mm 5mm,width=0.3\linewidth]{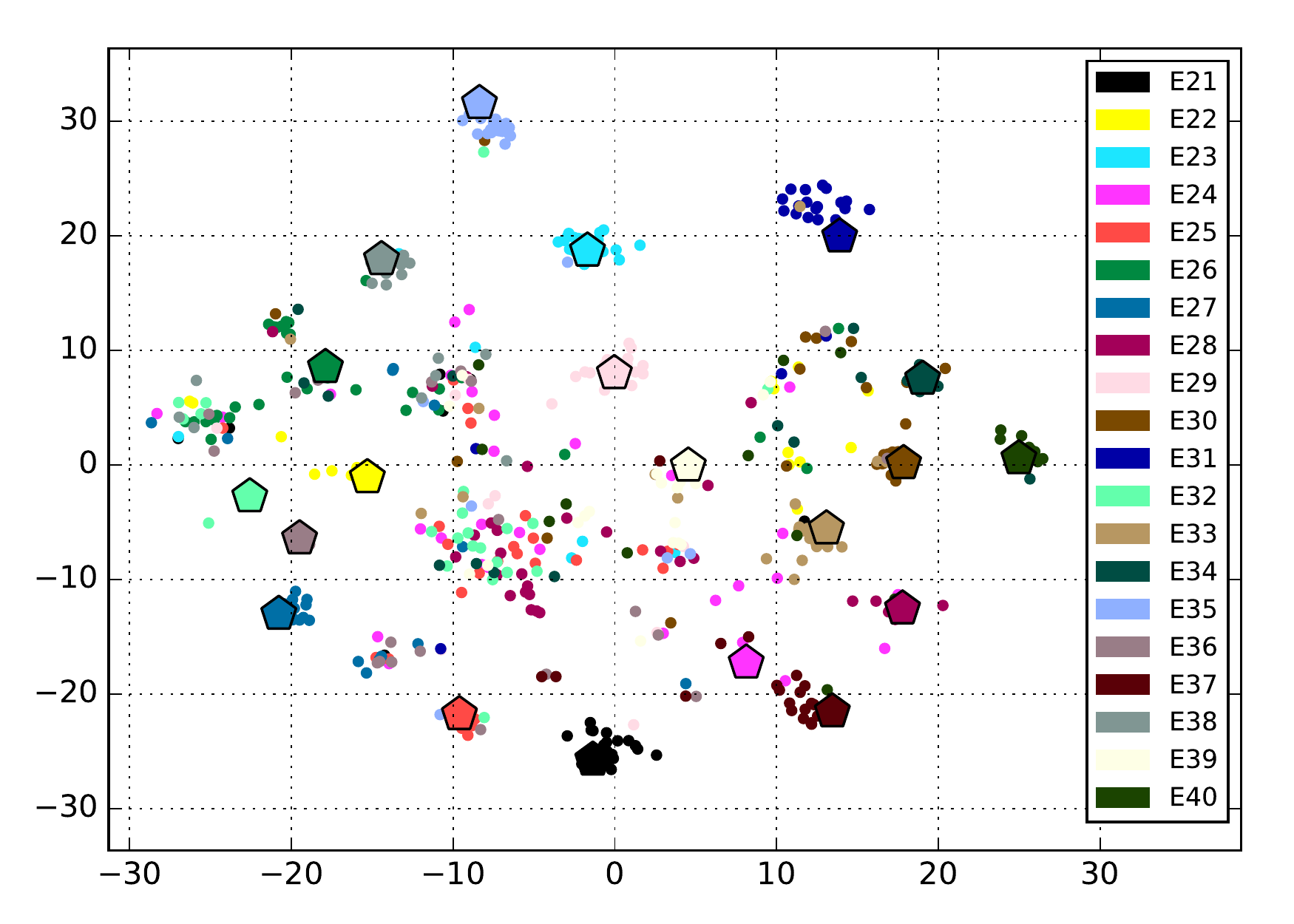}
\label{fig:4-5-6}}
\caption{We visualize the results of video embedding using the unified embedding model$^\mathcal{U}$ and baselines model$^\mathcal{V}$, model$^\mathcal{S}$. Each sub-figure shows how discriminant the representation of the embedded videos. Each dot represents a projected video, while each pentagon-shape represents a projected event description. We use t-SNE to visualize the result.}
\label{fig:4-5}
\end{figure*}



Next, we quantitatively demonstrate the benefit of the textual embedding $f_\mathcal{T}(\cdot)$. In contrast to the main model, see section~\ref{sec:model}, we investigate baseline \textbf{model}$^\mathcal{V}$, where we discard the textual embedding $f_\mathcal{T}(\cdot)$ and consider only the visual embedding $f_\mathcal{V}(\cdot)$. We project a video $v$ on the LSI representation $y$ of the related event $t$. Thus, this baseline falls in the first family of methods, see figure~\ref{fig:2-1}(a). It is optimized using mean-squared error (MSE) loss $\mathcal{L}^{\mathcal{V}}_{mse}$, see eq.~\ref{eq:loss_model_v}. The result of this baseline is reported in section~\ref{sec:results}, table~\ref{tbl:5-3}.

\begin{equation}
\mathcal{L}^{\mathcal{V}}_{mse} = \frac{1}{N} \sum_{i=1}^{N} \|y_i - f_\mathcal{V}(v_i; W_\mathcal{V})\|_2^2
\label{eq:loss_model_v}.
\end{equation}

Also, we train another baseline \textbf{model}$^\mathcal{C}$, which is similar to the aforementioned $^\mathcal{V}$ except instead of using MSE loss $\mathcal{L}^{\mathcal{V}}_{mse}$, see eq.~\eqref{eq:loss_model_v}, it uses contrastive loss $\mathcal{L}^{\mathcal{C}}_{con}$, as follows:

\begin{equation}
\begin{aligned}
\mathcal{L}^{\mathcal{C}}_{con} &= \frac{1}{2N} \sum_{i=1}^N h_i \cdot d_i^2 + (1-h_i) \max(1 - d_i, 0)^2,
\\
d_i &= \|y_i - f_\mathcal{V}(v_i; W_\mathcal{V})\|_2.
\label{eq:loss_model_c}
\end{aligned}
\end{equation}


\subsection{Unified Embedding and Metric Learning}
In this experiment, we demonstrate the benefit of the unified embedding. In contrast to our model presented in section~\ref{sec:model}, we investigate baseline \textbf{model}$^\mathcal{S}$, where this baseline does not learn joint embedding. Instead, it separately learns visual $f_\mathcal{V}(\cdot)$ and textual $f_\mathcal{T}(\cdot)$ projections. We model these projections as a shallow (2-layer) MLP trained to classify the data input into 500 event categories, using logistic loss, same as eq.~\eqref{eq:loss_logistic_u}.

We conduct another experiment to demonstrate the benefit of learning metric space. In contrast to our model presented in section~\ref{sec:model}, we investigate baseline \textbf{model}$^\mathcal{N}$, where we discard the metric learning layer. Consequently, this baseline learns the visual embedding is a shallow (2 layers) multi-layer perceptron with \texttt{tanh} non linearities. Also, we replace the contrastive loss $\mathcal{L}_c$, see eq.~\eqref{eq:loss_contrastive_u} with mean-squared error loss $\mathcal{L}_{mse}$, namely
\begin{equation}
\mathcal{L}^{\mathcal{N}}_{mse} = \frac{1}{N} \sum_{i=1}^{N} \|f_\mathcal{T}(t_i; W_\mathcal{T}) - f_\mathcal{V}(v_i; W_\mathcal{V})\|_2^2.
\label{eq:loss_model_n}
\end{equation}

During retrieval, this baseline embeds a test video $v_i$ and novel text query $t_i$ as features $z^v, z^t$ onto the common space $\mathcal{Z}$ using textual and visual embeddings $f_\mathcal{T}(\cdot), f_\mathcal{V}(\cdot)$, respectively. However, in a post-processing step, retrieval score $s_i$ for the video $v_i$ is the cosine distance between $\left( z^v, z^t \right)$. Similarly, all test videos are scored, ranked and retrieved. The results of the aforementioned baselines model$^\mathcal{S}$ and model$^\mathcal{N}$ are reported in table~\ref{tbl:5-3}.


\bigbreak
\partitle{Comparing Different Embeddings}. In the previous experiments, we investigated several baselines of the unified embedding (model$^\mathcal{U}$), namely visual-only embedding (model$^\mathcal{V}$), separate visual-textual embedding (model$^\mathcal{S}$) and non-metric visual-textual embedding (model$^\mathcal{N}$). In a qualitative manner, we compare the results of such embeddings. As shown in figure~\ref{fig:4-5}, we use these baselines to embed event videos of MED-13 and MED-14 datasets into the corresponding spaces. At the same time, we project the textual description of the events on the same space. Then, we use t-SNE~\cite{maaten2008visualizing} to visualize the result on 2D manifold. As seen, the unified embedding, see sub-figures~\ref{fig:4-5-5},~\ref{fig:4-5-6} learns more discriminant representations than the other baselines, see sub-figures~\ref{fig:4-5-1},~\ref{fig:4-5-2},~\ref{fig:4-5-3} and ~\ref{fig:4-5-4}. The same observation holds for both for MED-13 and MED-14 datasets.


\subsection{Mitigating Noise in EventNet}
Based of quantitative and qualitative analysis, we conclude that EventNet is noisy. Not only videos are unconstrained, but also some of the video samples are irrelevant to their event categories. EvenNet dataset~\cite{ye2015eventnet} is accompanied by 500-category CNN classifier. It achieves top-1 and top-5 accuracies of 30.67\% and 53.27\%, respectively. Since events in EventNet are structured as an ontological hierarchy, there is a total of 19 high-level categories. The classifier achieves top-1 and top-5 accuracies of 38.91\% and 57.67\%, respectively, over these high-level categories.

Based on these observations, we prune EventNet to remove noisy videos. To this end, first we represent each video as average pooling of ResNet \texttt{pool5} features. Then, we follow the conventional 5-fold cross validation with 5 rounds. For each round, we split the dataset into 5 subsets, 4 subsets $\mathcal{V}_{t}$ for training and the last $\mathcal{V}_{p}$ for pruning. Then we train a 2-layer MLP for classification. After training, we forward-pass the videos of $\mathcal{V}_{p}$ and rule-out the mis-classified ones.

The intuition behind pruning is that we rather learn salient event concepts using less video samples than learn noisy concepts with more samples. Pruning reduced the total number of videos by 26\%, from 90.2k to 66.7k. This pruned dataset is all what we use in our experiments.

\subsection{Latent Topics in LSI}
\begin{figure}[!ht]
\begin{center}
\includegraphics[trim=10mm 5mm 2mm 5mm,width=1.0\linewidth]{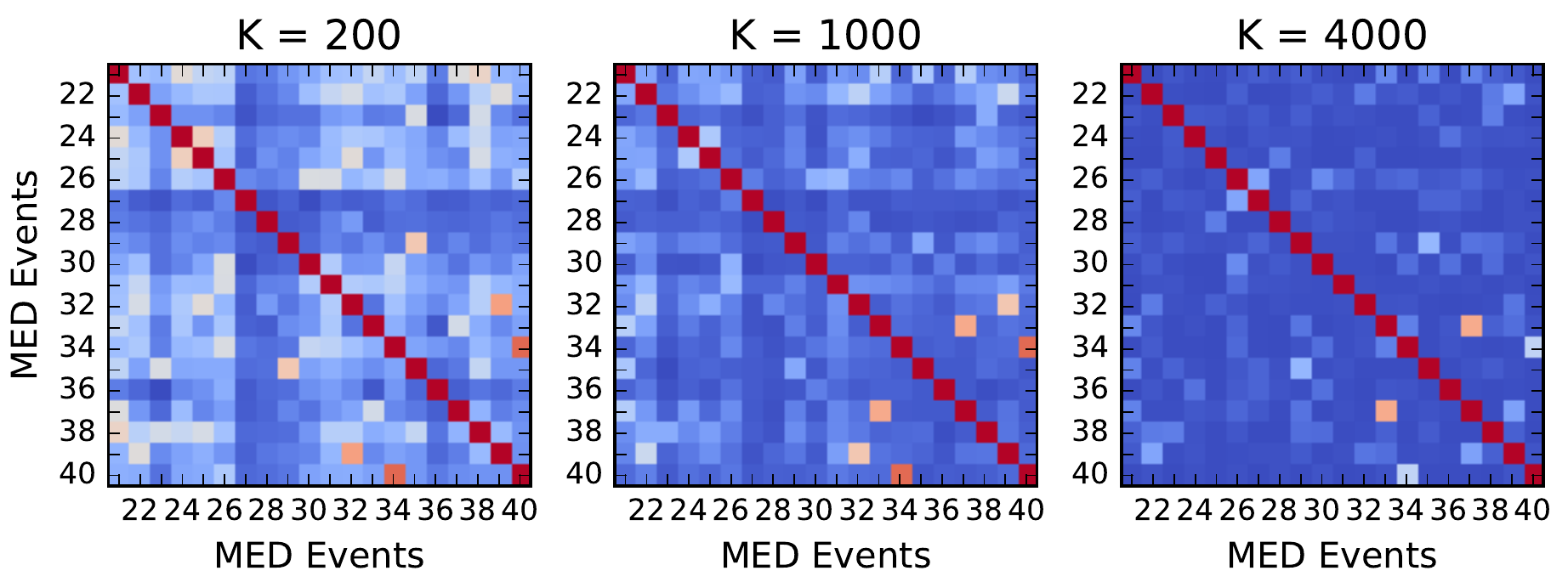}
\end{center}
\caption{Similarity matrix between LSI features of MED-14 events. The more the latent topics $(K)$ in LSI model, the higher the feature dimension, and the more discriminant the feature.}
\label{fig:4-6}
\end{figure}

When training LSI topic model on Wikipedia corpus, a crucial parameter is the number of latent topics $K$ the model constructs. We observe improvements in the performance directly proportional to increasing $K$. The main reason that the bigger the value of $K$, the more discriminant the LSI feature is. Figure~\ref{fig:4-6} confirms our understanding.

\section{Results}\label{sec:results}
\partitle{Evaluation metric}.
Since we are addressing, in essence, an information retrieval task, we rely on the average precision (AP) per event, and mean average precision (mAP) per dataset. We follow the standard evaluation method as in the relevant literature ~\cite{over2013trecvid,over2014trecvid,jiang2011consumer}.

\bigbreak
\partitle{Comparing against model baselines}.
In table~\ref{tbl:5-3}, we report the mAP score of our model baselines, previously discussed in the experiments, see section~\ref{sec:experiments}. The table clearly shows the marginal contribution of each of novelty for the proposed method.

\begin{table}[!ht]
\begin{center}
\setlength\tabcolsep{2pt}
\begin{tabular}{l|lcccc|cc}
\hline
Baseline & Loss & & Metric & $f_\mathcal{V}(\cdot)$ & $f_\mathcal{T}(\cdot)$ & MED13 & MED14 \\
\hline\hline
model$^\mathcal{V}$ & $\mathcal{L}_{mse}^{\mathcal{V}}$ & ~\eqref{eq:loss_model_v} & \xmark & \cmark & \xmark & 11.90 & 10.76 \\
model$^\mathcal{C}$ & $\mathcal{L}_{con}^{\mathcal{C}}$ & ~\eqref{eq:loss_model_c} & \cmark & \cmark & \xmark & 13.29 & 12.31 \\
model$^\mathcal{S}$ &  $\mathcal{L}_{log}$ & ~\eqref{eq:loss_logistic_u} & \xmark & \cmark& \cmark & 15.60 & 13.49 \\
model$^\mathcal{N}$ & $\mathcal{L}^{\mathcal{N}}_{mse}$ & ~\eqref{eq:loss_model_n} & \xmark & \cmark & \cmark & 15.92 & 14.36 \\
\hline\hline
model$^\mathcal{U}$ & $\mathcal{L}^{\mathcal{U}}$ & ~\eqref{eq:loss_model_u} & \cmark & \cmark & \cmark & 17.86 & 16.67 \\
\hline
\end{tabular}
\end{center}
\caption{Comparison between the unified embedding and other baselines. The unified embedding model$^\mathcal{U}$ achieves the best results on MED-13 and MED-14 datasets.}
\label{tbl:5-3}
\end{table}

\begin{figure*}[!ht]
\vspace*{\fill}
\centering
\subfigure[MED-13 Dataset]{
\includegraphics[trim=5mm 0mm 2mm 10mm,width=1.0\linewidth]{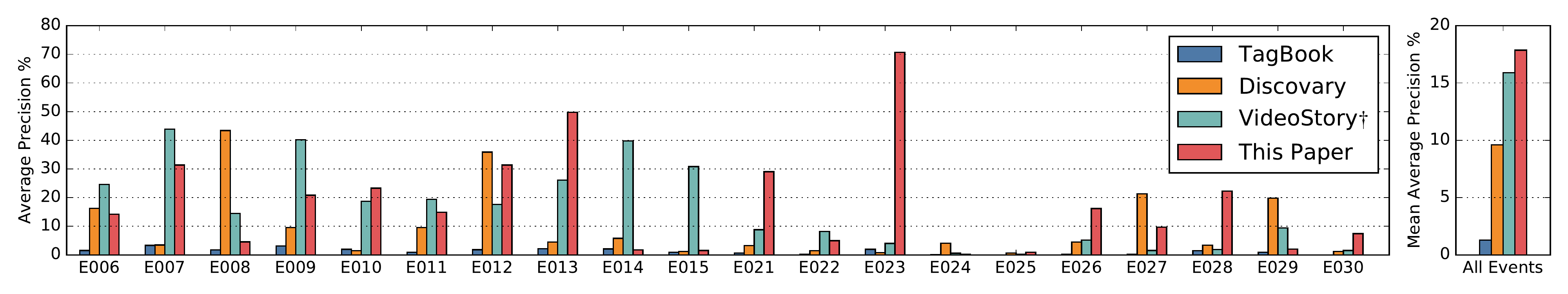}
\label{fig:5-1-1}}
\par\vfill
\subfigure[MED-14 Dataset]{
\includegraphics[trim=5mm 0mm 2mm 10mm,width=1.0\linewidth]{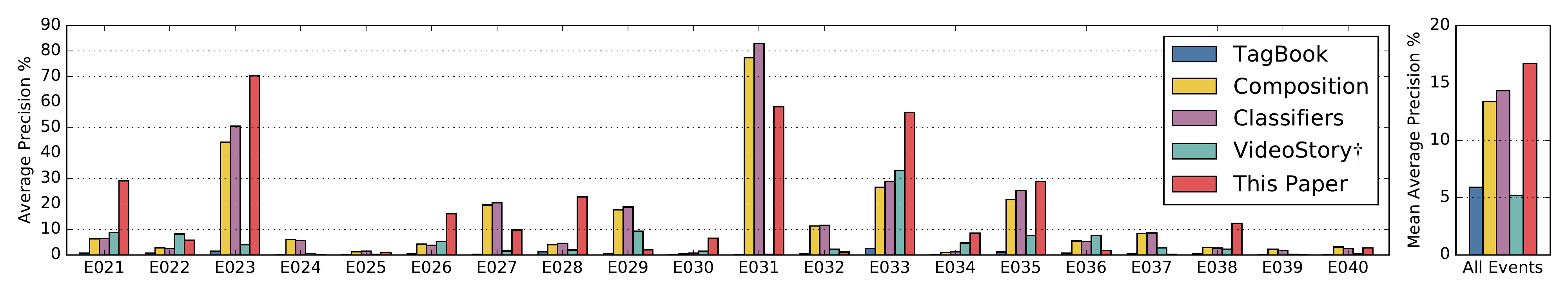}
\label{fig:5-1-2}}
\caption{Event detection accuracies: per-event average precision (AP\%) and per-dataset mean average precision (mAP\%) for MED-13 and MED-14 datasets. We compare our results against TagBook~\cite{mazloom2015tagbook}, Discovary~\cite{chang2015semantic}, Composition~\cite{chang2016dynamic}, Classifiers~\cite{changthey} and VideoStory~\cite{habibian2015videostory}.}
\label{fig:5-1}
\end{figure*}

\partitle{Comparing against related work}.
We report the performance of our method, the unified embedding model$^\mathcal{U}$ on TRECVID MED-13 and MED-14 datasets. When compared with the related works, our method improves over the state-of-the-art by a considerable margin, as shown in table~\ref{tbl:5-2} and figure~\ref{fig:5-1}.

\begin{table}[!ht]
\begin{center}
\setlength\tabcolsep{2pt}
\begin{tabular}{lr|cc}
\hline
Method & & MED13 & MED14 \\
\hline\hline
TagBook~\cite{mazloom2015tagbook} & ToM \textquotesingle15 & 12.90 & 05.90 \\
Discovary~\cite{chang2015semantic} & ICAI \textquotesingle15 & 09.60 & --  \\
Composition~\cite{chang2016dynamic} & AAAI \textquotesingle16 & 12.64 & 13.37 \\
Classifiers~\cite{changthey} & CVPR \textquotesingle16 & 13.46 & \underline{14.32} \\
VideoStory$^\dagger$~\cite{habibian2015videostory} & PAMI \textquotesingle16 & 15.90 & 05.20 \\
VideoStory$^*$~\cite{habibian2015videostory} & PAMI \textquotesingle16 & \textbf{20.00} & 08.00 \\
\hline
\hline
This Paper (model$^\mathcal{U}$) & & \underline{17.86} & \textbf{16.67} \\
\hline
\end{tabular}
\end{center}
\caption{Performance comparison between our model and related works. We report the mean average precision (mAP\%) for MED-13 and MED-14 datasets.}
\label{tbl:5-2}
\end{table}

It is important to point out that VideoStory$^\dagger$ uses only object feature representation, so its comparable to our method. However, VideoStory$^*$ uses motion feature representation and expert text query (i.e. using term-importance matrix $\bm{H}$ in ~\cite{habibian2015videostory}). To rule out the marginal effect of using different datasets and features, we train VideoStory and report results in table~\ref{tbl:5-1}. Clearly, CNN features and video exemplars in the training set can improve the model accuracy, but our method improves against VideoStory when trained on the same dataset and using the same features. Other works (Classifiers~\cite{changthey}, Composition~\cite{chang2016dynamic}) use both image and action concept classifiers. Nonetheless, our method improves over them using only object-centric CNN feature representations.

\begin{table}[!ht]
\begin{center}
\setlength\tabcolsep{2pt}
\begin{tabular}{l|ll|c}
\hline
Method & Training Set & CNN Feat. & MED14 \\
\hline\hline
VideoStory & VideoStory46k~\cite{habibian2015videostory} & GoogleNet & 08.00 \\
VideoStory & FCVID~\cite{jiang2017exploiting} & GoogleNet & 11.84 \\
VideoStory & EventNet~\cite{ye2015eventnet} & GoogleNet & 14.52 \\
VideoStory & EventNet~\cite{ye2015eventnet} & ResNet & 15.80 \\
\hline\hline
This Paper & EventNet~\cite{ye2015eventnet} & ResNet & 16.67 \\
\hline
\end{tabular}
\end{center}
\caption{Our method improves over VideoStory when trained on the same dataset and using the same feature representation.}
\label{tbl:5-1}
\end{table}

\section{Conclusion}\label{sec:conclusions}
In this paper, we presented a novel approach for detecting events in unconstrained web videos, in a zero-exemplar fashion. Rather than learning separate embeddings form cross-modal datasets, we proposed a unified embedding where several cross-modalities are jointly projected. This enables end-to-end learning. On top of this, we exploited the fact that zero-exemplar is posed as retrieval task and proposed to learn metric space. This enables measuring the similarities between the embedded modalities using this very space.

We experimented the novelties and demonstrated how they contribute to improving the performance. We complemented this by improvements over the state-of-the-art by considerable margin on MED-13 and MED-14 datasets.

However, the question still remains, how can we discriminate between these two MED events ``34: fixing musical instrument" and ``40: tuning musical instrument". We would like to argue that temporal modeling for human actions in videos is of absolute necessity to achieve such fine-grained event recognition. In future research, we would like to focus on human-object interaction in videos and how to model it temporally.

\subsection*{Acknowledgment}\label{acknowledgement}
We thank Dennis Koelma, Masoud Mazloom and Cees Snoek\footnote{\url{{kolema,m.mazloom,cgmsnoek}@uva.nl}} for lending their insights and technical support for this work.

{\small
\bibliographystyle{unsrt}
\bibliography{bib}

\begin{thebibliography}{10}

\bibitem{over2013trecvid}
Paul Over, George Awad, Jon Fiscus, Greg Sanders, and Barbara Shaw.
\newblock Trecvid 2013--an introduction to the goals, tasks, data, evaluation
  mechanisms, and metrics.
\newblock In {\em TRECVID Workshop}, 2013.

\bibitem{over2014trecvid}
Paul Over, Jon Fiscus, Greg Sanders, David Joy, Martial Michel, George Awad,
  Alan Smeaton, Wessel Kraaij, and Georges Qu{\'e}not.
\newblock Trecvid 2014--an overview of the goals, tasks, data, evaluation
  mechanisms and metrics.
\newblock In {\em TRECVID Workshop}, 2014.

\bibitem{jiang2015bridging}
Lu~Jiang, Shoou-I Yu, Deyu Meng, Teruko Mitamura, and Alexander~G Hauptmann.
\newblock Bridging the ultimate semantic gap: A semantic search engine for
  internet videos.
\newblock In {\em ICMR}, 2015.

\bibitem{habibian2014composite}
Amirhossein Habibian, Thomas Mensink, and Cees~GM Snoek.
\newblock Composite concept discovery for zero-shot video event detection.
\newblock In {\em ICMR}, 2014.

\bibitem{habibian2015discovering}
Amirhossein Habibian, Thomas Mensink, and Cees~GM Snoek.
\newblock Discovering semantic vocabularies for cross-media retrieval.
\newblock In {\em ICMR}, 2015.

\bibitem{mazloom2014conceptlets}
Masoud Mazloom, Efstrastios Gavves, and Cees G.~M. Snoek.
\newblock Conceptlets: Selective semantics for classifying video events.
\newblock In {\em IEEE TMM}, 2014.

\bibitem{chang2015semantic}
Xiaojun Chang, Yi~Yang, Alexander~G Hauptmann, Eric~P Xing, and Yao-Liang Yu.
\newblock Semantic concept discovery for large-scale zero-shot event detection.
\newblock In {\em IJCAI}, 2015.

\bibitem{chang2016dynamic}
Xiaojun Chang, Yi~Yang, Guodong Long, Chengqi Zhang, and Alexander~G Hauptmann.
\newblock Dynamic concept composition for zero-example event detection.
\newblock In {\em arXiv}, 2016.

\bibitem{changthey}
Xiaojun Chang, Yao-Liang Yu, Yi~Yang, and Eric~P Xing.
\newblock They are not equally reliable: Semantic event search using
  differentiated concept classifiers.
\newblock In {\em IEEE CVPR}, 2016.

\bibitem{lu2016zero}
Yi-Jie Lu.
\newblock Zero-example multimedia event detection and recounting with
  unsupervised evidence localization.
\newblock In {\em ACM MM}, 2016.

\bibitem{jiang2015fast}
Lu~Jiang, Shoou-I Yu, Deyu Meng, Yi~Yang, Teruko Mitamura, and Alexander~G
  Hauptmann.
\newblock Fast and accurate content-based semantic search in 100m internet
  videos.
\newblock In {\em ACM MM}, 2015.

\bibitem{mensink2014costa}
Thomas Mensink, Efstratios Gavves, and Cees Snoek.
\newblock Costa: Co-occurrence statistics for zero-shot classification.
\newblock In {\em IEEE CVPR}, 2014.

\bibitem{gavves2015activetransferlearning}
E.~Gavves, T.~E.~J. Mensink, T.~Tommasi, C.~G.~M. Snoek, and T~Tuytelaars.
\newblock Active transfer learning with zero-shot priors: Reusing past datasets
  for future tasks.
\newblock In {\em IEEE ICCV}, 2015.

\bibitem{ye2015eventnet}
Guangnan Ye, Yitong Li, Hongliang Xu, Dong Liu, and Shih-Fu Chang.
\newblock Eventnet: A large scale structured concept library for complex event
  detection in video.
\newblock In {\em ACM MM}, 2015.

\bibitem{wikihow}
Wikihow.
\newblock \url{http://wikihow.com}.

\bibitem{habibian2014videostory}
Amirhossein Habibian, Thomas Mensink, and Cees~GM Snoek.
\newblock Videostory: A new multimedia embedding for few-example recognition
  and translation of events.
\newblock In {\em ACM MM}, 2014.

\bibitem{habibian2015videostory}
Amirhossein Habibian, Thomas Mensink, and Cees~GM Snoek.
\newblock Videostory embeddings recognize events when examples are scarce.
\newblock In {\em IEEE TPAMI}, 2016.

\bibitem{mazloom2015tagbook}
Masoud Mazloom, Xirong Li, and Cees Snoek.
\newblock Tagbook: A semantic video representation without supervision for
  event detection.
\newblock In {\em IEEE TMM}, 2015.

\bibitem{wu2014zero}
Shuang Wu, Sravanthi Bondugula, Florian Luisier, Xiaodan Zhuang, and Pradeep
  Natarajan.
\newblock Zero-shot event detection using multi-modal fusion of weakly
  supervised concepts.
\newblock In {\em IEEE CVPR}, 2014.

\bibitem{elhoseiny2015zero}
Mohamed Elhoseiny, Jingen Liu, Hui Cheng, Harpreet Sawhney, and Ahmed Elgammal.
\newblock Zero-shot event detection by multimodal distributional semantic
  embedding of videos.
\newblock In {\em arXiv}, 2015.

\bibitem{mikolov2013exploiting}
Tomas Mikolov, Quoc~V Le, and Ilya Sutskever.
\newblock Exploiting similarities among languages for machine translation.
\newblock In {\em arXiv}, 2013.

\bibitem{mikolov2013distributed}
Tomas Mikolov, Ilya Sutskever, Kai Chen, Greg~S Corrado, and Jeff Dean.
\newblock Distributed representations of words and phrases and their
  compositionality.
\newblock In {\em NIPS}, 2013.

\bibitem{jiang2014easy}
Lu~Jiang, Deyu Meng, Teruko Mitamura, and Alexander~G Hauptmann.
\newblock Easy samples first: Self-paced reranking for zero-example multimedia
  search.
\newblock In {\em ACM MM}, 2014.

\bibitem{jiang2014zero}
Lu~Jiang, Teruko Mitamura, Shoou-I Yu, and Alexander~G Hauptmann.
\newblock Zero-example event search using multimodal pseudo relevance feedback.
\newblock In {\em ICMR}, 2014.

\bibitem{agharwal2016tag}
Arnav Agharwal, Rama Kovvuri, Ram Nevatia, and Cees~GM Snoek.
\newblock Tag-based video retrieval by embedding semantic content in a
  continuous word space.
\newblock In {\em IEEE WACV}, 2016.

\bibitem{tran2014c3d}
Du~Tran, Lubomir Bourdev, Rob Fergus, Lorenzo Torresani, and Manohar Paluri.
\newblock C3d: generic features for video analysis.
\newblock In {\em arXiv}, 2014.

\bibitem{simonyan2014two}
Karen Simonyan and Andrew Zisserman.
\newblock Two-stream convolutional networks for action recognition in videos.
\newblock In {\em NIPS}, 2014.

\bibitem{wang2011action}
Heng Wang, Alexander Kl{\"a}ser, Cordelia Schmid, and Cheng-Lin Liu.
\newblock Action recognition by dense trajectories.
\newblock In {\em IEEE CVPR}, 2011.

\bibitem{uijlings2015video}
J~Uijlings, IC~Duta, Enver Sangineto, and Nicu Sebe.
\newblock Video classification with densely extracted hog/hof/mbh features: an
  evaluation of the accuracy/computational efficiency trade-off.
\newblock In {\em IJMIR}, 2015.

\bibitem{fernandoTPAMIrankpooling}
B.~Fernando, E.~Gavves, J.~Oramas, A.~Ghodrati, and T.~Tuytelaars.
\newblock Rank pooling for action recognition.
\newblock In {\em IEEE TPAMI}, 2016.

\bibitem{muda2010voice}
Lindasalwa Muda, Mumtaj Begam, and I~Elamvazuthi.
\newblock Voice recognition algorithms using mel frequency cepstral coefficient
  (mfcc) and dynamic time warping (dtw) techniques.
\newblock In {\em arXiv}, 2010.

\bibitem{kumar2016audio}
Anurag Kumar and Bhiksha Raj.
\newblock Audio event detection using weakly labeled data.
\newblock In {\em arXiv}, 2016.

\bibitem{jing2016discriminative}
Liping Jing, Bo~Liu, Jaeyoung Choi, Adam Janin, Julia Bernd, Michael~W Mahoney,
  and Gerald Friedland.
\newblock A discriminative and compact audio representation for event
  detection.
\newblock In {\em ACM MM}, 2016.

\bibitem{lecun1998gradient}
Yann LeCun, L{\'e}on Bottou, Yoshua Bengio, and Patrick Haffner.
\newblock Gradient-based learning applied to document recognition.
\newblock In {\em IEEE}.

\bibitem{he2015deep}
Kaiming He, Xiangyu Zhang, Shaoqing Ren, and Jian Sun.
\newblock Deep residual learning for image recognition.
\newblock In {\em arXiv}, 2015.

\bibitem{krizhevsky2012imagenet}
Alex Krizhevsky, Ilya Sutskever, and Geoffrey~E Hinton.
\newblock Imagenet classification with deep convolutional neural networks.
\newblock In {\em NIPS}, 2012.

\bibitem{simonyan2014very}
Karen Simonyan and Andrew Zisserman.
\newblock Very deep convolutional networks for large-scale image recognition.
\newblock In {\em arXiv}, 2014.

\bibitem{gan2016you}
Chuang Gan, Ting Yao, Kuiyuan Yang, Yi~Yang, and Tao Mei.
\newblock You lead, we exceed: Labor-free video concept learning by jointly
  exploiting web videos and images.
\newblock In {\em IEEE CVPR}, 2016.

\bibitem{simonyan2013fisher}
Karen Simonyan, Omkar~M Parkhi, Andrea Vedaldi, and Andrew Zisserman.
\newblock Fisher vector faces in the wild.
\newblock In {\em BMVC}, 2013.

\bibitem{arandjelovic2013all}
Relja Arandjelovic and Andrew Zisserman.
\newblock All about vlad.
\newblock In {\em IEEE CVPR}, 2013.

\bibitem{mettes2015bag}
Pascal Mettes, Jan~C van Gemert, Spencer Cappallo, Thomas Mensink, and Cees~GM
  Snoek.
\newblock Bag-of-fragments: Selecting and encoding video fragments for event
  detection and recounting.
\newblock In {\em ICMR}, 2015.

\bibitem{sutskever2014sequence}
Ilya Sutskever, Oriol Vinyals, and Quoc~V Le.
\newblock Sequence to sequence learning with neural networks.
\newblock In {\em NIPS}, 2014.

\bibitem{le2014distributed}
Quoc~V Le and Tomas Mikolov.
\newblock Distributed representations of sentences and documents.
\newblock In {\em ICML}, 2014.

\bibitem{blei2003latent}
David~M Blei, Andrew~Y Ng, and Michael~I Jordan.
\newblock Latent dirichlet allocation.
\newblock In {\em JMLR}, 2003.

\bibitem{deerwester1990indexing}
Scott Deerwester, Susan~T Dumais, George~W Furnas, Thomas~K Landauer, and
  Richard Harshman.
\newblock Indexing by latent semantic analysis.
\newblock In {\em JACS}, 1990.

\bibitem{chopra2005learning}
Sumit Chopra, Raia Hadsell, and Yann LeCun.
\newblock Learning a similarity metric discriminatively, with application to
  face verification.
\newblock In {\em IEEE CVPR}, 2005.

\bibitem{szegedy2015going}
Christian Szegedy, Wei Liu, Yangqing Jia, Pierre Sermanet, Scott Reed, Dragomir
  Anguelov, Dumitru Erhan, Vincent Vanhoucke, and Andrew Rabinovich.
\newblock Going deeper with convolutions.
\newblock In {\em IEEE CVPR}, 2015.

\bibitem{zhou2016places}
Bolei Zhou, Aditya Khosla, Agata Lapedriza, Antonio Torralba, and Aude Oliva.
\newblock Places: An image database for deep scene understanding.
\newblock In {\em arXiv}, 2016.

\bibitem{wikipedia}
Wikipedia, 2016.
\newblock \url{http://wikipedia.com}.

\bibitem{kiros2015skip}
Ryan Kiros, Yukun Zhu, Ruslan~R Salakhutdinov, Richard Zemel, Raquel Urtasun,
  Antonio Torralba, and Sanja Fidler.
\newblock Skip-thought vectors.
\newblock In {\em NIPS}, 2015.

\bibitem{maaten2008visualizing}
Laurens van~der Maaten and Geoffrey Hinton.
\newblock Visualizing data using t-sne.
\newblock In {\em JMLR}, 2008.

\bibitem{jiang2011consumer}
Yu-Gang Jiang, Guangnan Ye, Shih-Fu Chang, Daniel Ellis, and Alexander~C Loui.
\newblock Consumer video understanding: A benchmark database and an evaluation
  of human and machine performance.
\newblock In {\em ICMR}, 2011.

\bibitem{jiang2017exploiting}
Yu-Gang Jiang, Zuxuan Wu, Jun Wang, Xiangyang Xue, and Shih-Fu Chang.
\newblock Exploiting feature and class relationships in video categorization
  with regularized deep neural networks.
\newblock In {\em IEEE TPAMI}, 2017.

\end{thebibliography}
}

\end{document}